\pdfoutput=1
\documentclass[11pt]{article}

    \usepackage[preprint]{acl}

\usepackage{times}
\usepackage{latexsym}

\usepackage[T1]{fontenc}
\usepackage[utf8]{inputenc}

\usepackage{microtype}

\usepackage{inconsolata}

\usepackage{graphicx}
\usepackage{booktabs}       % for \toprule etc.
\usepackage{tabularx}       % for column type X
\usepackage{enumitem}
\usepackage{placeins}
\usepackage{multicol}
\usepackage{caption}
\usepackage[export]{adjustbox}

\usepackage{subfig}

\usepackage{booktabs}
\usepackage{hyperref}
\usepackage{longtable}

\usepackage{bbm}
\usepackage{stmaryrd}
\usepackage{amsmath,amssymb}

\usepackage{amsthm}

\usepackage{refcount}
\usepackage{fancyvrb}

\newcommand{\package}[0]{\texttt{CLEAR}}

\usepackage{booktabs}
\usepackage{multirow}
\usepackage{hyperref}
\usepackage{listings}
\usepackage[most]{tcolorbox}
\usepackage{xcolor}

\definecolor{myred}{HTML}{EC6460}   % Replace FF0000 with your desired HTML code for red
\definecolor{myyellow}{HTML}{ECBD60  } % Replace FFFF00 with your desired HTML code for yellow
\definecolor{mygreen}{HTML}{90EC60}  % Replace 008000 with your desired HTML code for green
\definecolor{mygray}{HTML}{AEB0AC}   % Replace 808080 with your desired HTML code for gray#3c3273
\definecolor{deeppurple}{HTML}{3c3273}
\definecolor{mygray}{RGB}{240,240,240}
\definecolor{darkgray}{RGB}{64,64,64} % This is a dark gray. Adjust the numbers to change the shade.

\newtcolorbox{mycodebox}{
    enhanced,
    fonttitle=\bfseries,
    colback=white,
    colframe=mygray,
    colbacktitle=mygray,
    coltitle=black,
    rounded corners,
    boxrule=0.5mm,
    top=5mm,
    interior style={top color=mygray, bottom color=white},
    overlay={
        \node[anchor=west, font=\bfseries] at ([xshift=5pt, yshift=-10pt]frame.north west)
        {\textcolor{myred}{$\bullet$} \textcolor{myyellow}{$\bullet$} \textcolor{mygreen}{$\bullet$}};
    },
    left=4mm,
    right=2mm,
    bottom=2mm,
}

\newtcolorbox{myshellbox}{
    enhanced,
    fonttitle=\bfseries,
    colback=white,
    colframe=mygray,
    coltext=white,
    colbacktitle=mygray,
    coltitle=black,
    rounded corners,
    boxrule=0.5mm,
    top=5mm,
    interior style={top color=black, bottom color=darkgray},
    overlay={
        \node[anchor=west, font=\bfseries] at ([xshift=5pt, yshift=-10pt]frame.north west)
        {\textcolor{myred}{$\bullet$} \textcolor{myyellow}{$\bullet$} \textcolor{mygreen}{$\bullet$}};
    },
    left=2mm,
    right=2mm,
    bottom=2mm,
}

\usepackage[utf8]{inputenc}
\usepackage{tcolorbox}
\usepackage{fvextra}        % Enables \VerbatimInput with breaklines
\usepackage{caption}
\usepackage{graphicx} 
\usepackage{dashrule}

\title{\package{}: Error Analysis via LLM-as-a-Judge Made Easy}

\author{
  \textbf{Asaf Yehudai\textsuperscript{I,H}}\thanks{Equal contribution.},
  \textbf{Lilach Eden\textsuperscript{I}}\footnotemark[1],
  \textbf{Yotam Perlitz\textsuperscript{I}},
\\
  \textbf{Roy Bar-Haim\textsuperscript{I}},
  \textbf{Michal Shmueli-Scheuer\textsuperscript{I}}
\\
\\
  \textsuperscript{I}IBM Research
  \textsuperscript{H}The Hebrew University of Jerusalem
\\
\{Asaf.Yehudai, Y.Perlitz\}@ibm.com
\{lilache, roybar, shmueli\}@il.ibm.com \\
}

\begin{document}
\maketitle
\begin{abstract}
The evaluation of Large Language Models (LLMs) increasingly relies on other LLMs acting as judges. However, current evaluation paradigms typically yield a single score or ranking, answering \textit{which} model is better but not \textit{why}. While essential for benchmarking, these top-level scores obscure the specific, actionable reasons behind a model's performance.
% This lack of explanatory power hinders developers in understanding and improving their models.
To bridge this gap, we introduce \package{}, an interactive, open-source package for LLM-based error analysis.
% \benchmark{} compose of two steps, first, it g
\package{} first generates per-instance textual feedback, then it creates a set of system-level error issues, and quantifies the prevalence of each identified issue.
% summarizes these into a concise set of system-level critiques, and finally quantifies the prevalence of each identified issue.
Our package also provides users with an interactive dashboard that allows for a comprehensive error analysis 
through aggregate visualizations, applies interactive filters to isolate specific issues or score ranges, and drills down to the individual instances that exemplify a particular behavioral pattern. 
We demonstrate \package{} analysis for RAG and Math benchmarks, and showcase its utility through a user case study.

% \url{https://ibm.biz/CLEAR-demo-video}
% \newline
Code: \url{https://ibm.biz/CLEAR-code-repo}

% empowering developers with the concrete, evidence-backed insights needed for targeted model improvement.

% Our tool provides interpretable and actionable insights, enabling users to diagnose model weaknesses and guide improvements.

\end{abstract}

% Placeholder commands for consistency with the original paper's style

\section{Introduction}

The evaluation of generative AI systems is rapidly adopting the LLM-as-a-Judge (LLMaJ) paradigm \cite{zheng2023llmaaj}, where automatic evaluations by LLMs complement or even replace human annotators. LLM-based judges are commonly applied to rate or score the quality of LLM responses \cite{liu2023gevalnlgevaluationusing}, or to choose a preferred response out of multiple candidates. 

Aggregating these judgments over many examples provides AI developers with a robust assessment of their system, as well as systematic comparison and ranking of different systems or models \cite{gera2025justrankbenchmarkingllmjudges}. However, these scores or ratings alone provide little insight into the model's behavior. AI developers still rely on tedious, manual error analysis to identify recurring issues, understand the current limitations of their system and effectively plan and prioritize the next iteration of improvements.    

%A developer might learn their model is ranked lower than a competitor, but they are left to speculate about the specific, systematic reasons for %the performance gap. Are its responses too verbose? Does it frequently refuse to answer certain types of questions? Does it struggle with following complex instructions?    

%LLM-based judges are increasingly relied upon to score responses \cite{liu2023gevalnlgevaluationusing}, rank models \cite{gera2025justrankbenchmarkingllmjudges}, and assess the overall quality of a given system.

%This paradigm has proven effective for creating leaderboards \cite{dubois2025lengthcontrolledalpacaevalsimpleway} and making comparative judgments \cite{lambert2024rewardbenchevaluatingrewardmodels}. These evaluations typically aggregate instance-level scores from pairwise comparisons or direct scoring into a single system-level metric, such as a win-rate or an average score. While these metrics are valuable for summative assessment, they offer little insight into the underlying reasons for a model's performance. A developer might learn their model is ranked lower than a competitor, but they are left to speculate about the specific, systematic reasons for the performance gap. Are its responses too verbose? Does it frequently refuse to answer certain types of questions? Does it struggle with following complex instructions?

In this work, we introduce \package{}, a novel interactive tool for AI developers, designed to reduce the overhead of manual error analysis.
Our approach utilizes an LLMaJ for generating textual feedback, and conducts discovery of recurring issues via Key Points Analysis (KPA) \cite{bar-haim-etal-2020-arguments}.
% Our approach utilizes an LLMaJ for generating textual feedback, and conducts recurrent issues discovery via Key Points Analysis (KPA) \cite{bar-haim-etal-2020-arguments}.
This method allows us to provide structured, textual feedback that characterizes and quantifies a model's recurring weaknesses and issues across a whole dataset. These insights may guide further improvements, such as prompt engineering, model fine-tuning, or choosing a different LLM.

% (\packagelong{})

\begin{figure*}[t!]
    \centering
    \includegraphics[width=\linewidth,trim={0 40pt 20pt 0}, ]
    {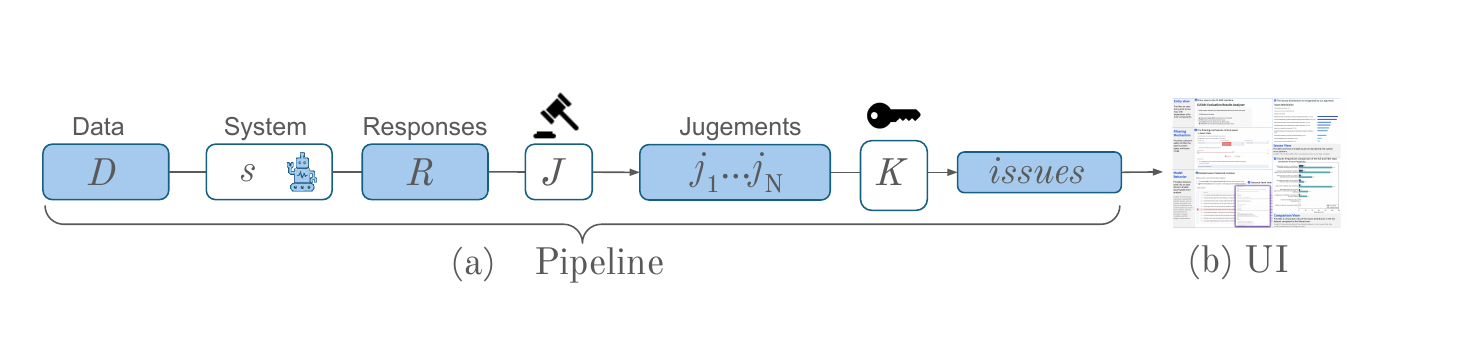} 
    \caption{\textbf{The \package{} Framework.} (a) Pipeline- Given a dataset ($D$) and a target system ($s$), the system generates responses ($R$). A judge ($J$) provides per-instance textual feedback and a score ($\{j_i\}_{i=1}^N$). A Key Point Analysis module ($K$) extracts recurring issues and maps them to the individual $j_i$'s. The discovered issues can be explored via the UI (b).}
    \label{fig:schema} \vspace{-4pt}
\end{figure*}

The \package{} pipeline is illustrated in Figure~\ref{fig:schema}. It starts with per-instance judgments, which include both a numeric score and textual feedback. It then employs a KPA module to categorize these individual critiques into a concise set of automatically-discovered issues. Each identified issue is mapped back to its matching judgments, which provides quantification for its prevalence, and allows the user to drill down from an issue to its specific examples. Lastly, we provide a user interface that allows for easy and dynamic exploration of issues within the data. 

% \begin{figure*}[t!]
%     \centering
%     \includegraphics[width=\linewidth,
%         trim={0 20pt 0 0}, 
%         clip
%     ]
%     {Figures/Figure1_v3.pdf} 
%     \caption{\textbf{The \package{} Framework.} Given a dataset ($D$) and a target system ($s$), the system generates responses ($R$). A judge ($J$) provides per-instance textual feedback and a score ($\{j_i\}_{i=1}^N$). A Key Point Analysis module ($K$) extracts recurring issues and maps them to the individual $j_i$'s. The discovered issues can be explored via the UI.}
%     \label{fig:schema}
% \end{figure*}

To demonstrate our system's capabilities, we ran our system on several RAG (contextual question-answering) and math benchmarks. We analyzed responses from several systems using different LLM Judges and KPA implementations. 
%In \S\ref{} we show a few of the extracted key points. 
To further demonstrate our system usability for real-world AI developers, we also conducted a user study. Its results confirm the usefulness of \package{}, and its potential value for reducing the time and effort required for error analysis.

Our work makes the following contributions:
\begin{enumerate}
    \item We propose a novel setup for generating automated system-level issues by summarizing and structuring instance-level feedback.
    \item We present \package{}, an open-source demo tool that implements our proposed approach, and an interactive UI, which together provide an accessible way for researchers and developers to gain deeper insights into their models' behaviors.
    \item We demonstrate the system on multiple domains, and conduct a user study to confirm its effectiveness.
\end{enumerate}

\section{Method}
% A mathematical description of the method. D -> LLM -> J -> M -> Issues  --- Two implementation

\package{} is designed to produce system-level feedback by analyzing a model's behavior across a dataset. The full setup, illustrated in Figure~\ref{fig:schema}, takes as input a dataset of instructions and a target system, and outputs a concise, structured, and quantified summary of the system's recurring issues.

Formally, we assume a dataset $\mathcal{D} = \{x_n\}_{n=1}^N$ consisting of $N$ instructions, and a target system $s$. The system generates a corresponding set of responses $\mathcal{R} = \{r_n\}_{n=1}^N$, where each $r_n = s(x_n)$. Our framework then proceeds in two primary stages:

% \paragraph{Per-Instance Judgement.} 
An LLM-based judge $J$ is prompted to evaluate each pair $(x_n, r_n)$. For each instance, $J$ returns a tuple $j_n = (t_n, s_n)$, where $t_n$ is a natural language critique and $s_n$ is a numeric quality score. These instance-level judgments capture localized failures or strengths observed by the judge. We note that our setup is reference-less, yet when a reference is available, it can be used as context for the judge.

% \paragraph{2. Critique Aggregation via Key Point Analysis.} 
The second stage clusters recurring patterns across the textual feedbacks $\{t_n\}_{n=1}^N$ into a set of concise, interpretable issues $\{i_m\}_{m=1}^M$. For efficiency, and because the focus is on identifying shortcomings, only feedback associated with $s_n<1$ is considered during issue generation. Each $t_n$ is then linked to one or more relevant issues from this set. We explore two distinct implementations for this aggregation module, denoted $K$:

\textbf{Key Point Analysis (KPA)}: We adopt a classical KPA pipeline \cite{bar-haim-etal-2020-arguments, bar-haim-etal-2020-quantitative}, which is well-suited for texts containing short sentences, such as arguments or product reviews. To improve compatibility, we first break down each $t_n$ into a brief and well-formed sentence using an LLM. We then apply the KPA method to cluster the sentences and construct a set of issues over them. This clustering allows for mapping each judgment to the issues its sentences express.

\textbf{LLM-Based KPA}: As an alternative approach, we propose LLM-Based KPA. We start by summarizing each critique $t_n$ into a shorter, normalized form via an LLM call. We then prompt an LLM with a batch of these summaries to identify high-level recurring issues, and again to remove duplication and consolidate the final lists of issues. Finally, each $t_n$ is mapped to the derived issue set via a matching prompt. This process requires $\sim2N$ LLM calls. Implementation details are provided in Appendix~\ref{app:promts}.
% \mss{i suggest we add here a short cost analysis (as we did in the slides.).}

% For example, $t_j$ could be "The response is factually incorrect and hallucinates a source" or "The response follows all instructions but is overly verbose."

% \paragraph{2. Critique Summarization.}
% In the next step, denoted as $K$, we first want to extract from the $t_n$'s recurring issues $\{i_m\}_{m=1}^M$, and then map each $t_n$ to the relevant issues it exhibits. In this work, we examine two different implementations of this idea. The first is based on the Key-point analysis method \cite{bar-haim-etal-2020-arguments, bar-haim-etal-2020-quantitative}. The second utilizes LLM prompting to achieve the same goal. 

% Key Point Analysis (KPA) is designed to compress a corpus of texts into a group of concise key points. The analysis is designed for texts with relatively short sentences, such as arguments or product reviews. To better support this analysis, we prompt an LLM to rephrase the raw judgment strings into paragraphs of shorter sentences. We then apply to those sentences KPA tool to generate recurring issues and a mapping between the judgments and the issues.

% For the LLM-based implementation, we start by summarizing each one of the $N$ textual feedback in $\{t_1\}_{n=1}^{N}$. Then we prompt an LLM to extract key issues from the summaries when a large number of them are provided in the context. Based on the resultant issues, we map the judgment summaries back into the issues. Full prompts are in App. \ref{app:promts}.

% \begin{comment}
\begin{figure*}[ht!]
    \begin{center}        
    % Trim format: trim={left bottom right top}
    % Here we are trimming 0 from left, 150pt from bottom, 0 from right, 0 from top.
    \includegraphics[
        width=1\linewidth, 
        trim={0 100pt 0 0}, 
       clip
    ]{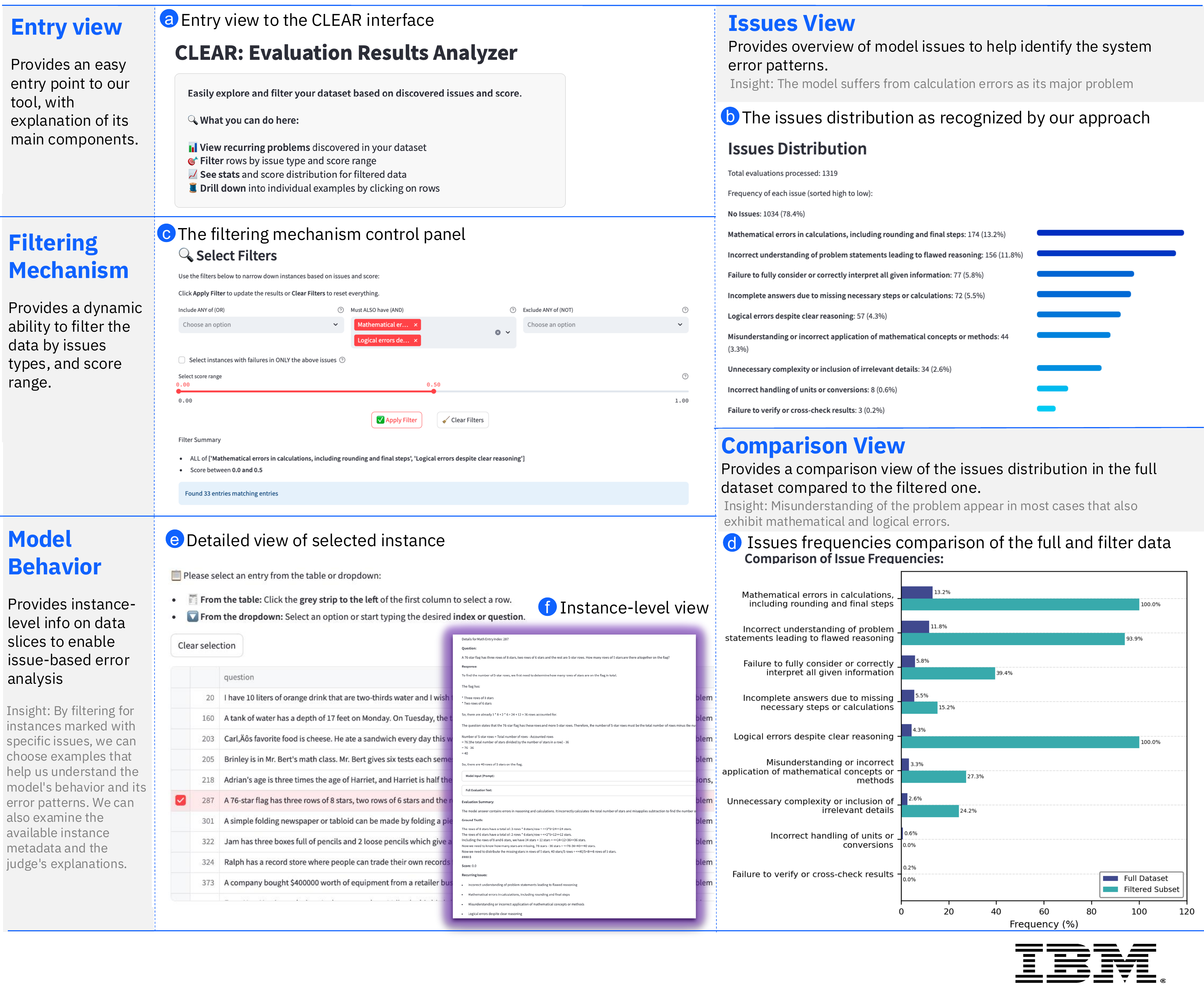} 
    \caption{
    % LM-Critique user interface
    The figure presents the key components of the \package{} tool for analyzing model evaluation results. (a) Entry point to the interface for exploring the model results and issues. (b) Issues View visualizes the distribution of detected model errors. (c) The Filtering Mechanism allows filtering based on issue types and scores to isolate relevant examples. (d) Comparison View contrasts issue frequencies between the full dataset and filtered subsets, highlighting co-occurrence patterns. (e/f) Model Behavior and Instance-Level View offer detailed, example-level insights to facilitate fine-grained error analysis and model diagnosis.
    \label{fig:ui}} %\vspace{-4pt}
    \end{center}
\end{figure*}
% \end{comment}

\section{\package{} Framework}

\subsection{Pipeline}
% To make our method easy to use and readily available, we provide \package{} implementation as a Python package. Our package supports an end-to-end solution, from generating responses and assessing the responses with a judge model, to applying a key point analysis component to derive issues. We design our system to be configurable by allowing to use of each part separately or any specific combination. We also need to modify the pipeline default prompts to address the specific task.

% In our default solution, the discovery of issues depends on the system, judge, and key point analysis component. Yet, some users may want a strict set of issues, while others will want a combination of a predefined set of issues together with dynamically discovered ones. To address those users' needs, we provide an adequate modification that can adjust our pipeline to their advantage.

To support easy integration and usability, we provide \package{} as a Python package available on PyPI. The package implements an end-to-end workflow: it generates model responses, evaluates them using an LLM judge, and performs key point analysis to identify recurring issues. Each component in the pipeline can be used independently or in combination, allowing users to customize the workflow to their specific needs or preferences.

% Moreover, we design the \package{} pipeline to be configurable. We provide support for different inference APIs providers. We support running evaluations on results obtained independently. We also allow customizing the default prompts used by the judge, tailoring them to specific tasks or use cases.
Moreover, the \package{} pipeline is designed to be highly configurable. It supports multiple inference API providers for generating predictions or, alternatively, allows running the evaluation step independently if predictions are already available. In cases where judgments have been obtained separately, the pipeline can directly execute the final evaluation step ($K$ step). 

% We also allow customizing the default prompts used by the judge, tailoring them to specific tasks or use cases.

% The pipeline is highly configurable. For example, users can choose to perform generation using the available\mss{maybe supported? it somehow gives the impression that LLMs are part of the pipeline, and i think we want to say that we provide the option to support different inference APIs } LLMs within the pipeline or run evaluations on results obtained from experiments conducted elsewhere. They can also customize the default prompts used by the judge, tailoring them to specific tasks or use cases.

%The pipeline is highly configurable. For instance, users can choose to modify the default prompts used in each module to better suit the target task. 
% While our default configuration discovers issues dynamically based on the system, judge, and dataset, some users may prefer a static set of predefined issues or a hybrid approach. %To support these variations, \package{} allows users to supply prior issue lists, combine them with discovered issues, or constrain the pipeline issues accordingly.
% To support these variations, \package{} allows users to either provide a predefined list of issues to be mapped directly to instances or supply the issues as evaluation criteria, guiding the judge to look for them explicitly while remaining open to discovering additional issues.

\paragraph{Evaluation Modes.}
To accommodate different preferences for issue discovery, \package{} supports three evaluation modes:
\textbf{(1) General}: issues are discovered dynamically using a general-purpose evaluation prompt, enabling broad, exploratory assessment without requiring data-specific prior knowledge; \textbf{(2) Task-specific}: users provide specific issues as evaluation criteria, guiding the judge while allowing for additional discoveries, and \textbf{(3) Static}  a predefined list of issues supplied by the user is given to the judge as the sole evaluation criteria and mapped directly to evaluation texts, without any dynamic discovery.

%By supporting these modes, \package{} offers flexibility ranging from entirely manual specification to fully automatic issue discovery, or a hybrid of both.

% Dynamic key-points - you can use your own, you can let the system discover them, or include them as prior by providing them to the judge.

\paragraph{Code Example}
% About code - include only to show it is very easy to use
% Here we provide a simple usage example. of a 
% Bash code:

\package{} can be installed via PyPI:

\begin{myshellbox}
\begin{lstlisting}
 $ pip install clear-eval
\end{lstlisting}
\end{myshellbox}

% Once installed, the full analysis can be executed with  three lines of Python code, using the following code snippet:

Once installed, the full analysis can be executed with  a single CLI command:
% , using the following code snippet:

%the full pipeline can be executed with just a few lines of code. The following snippet shows how to run the analysis pipeline on a dataset. % A complete working example is provided in Appendix~\ref{app:code}.

% To utilize \package{}, import the \package{} pipeline. This enables running the full pipeline over the input data. Here is a simple code example of running \package{} pipeline. In App. \S\ref{}, we provide a complete code example.

\begin{myshellbox}
\begin{lstlisting}
$ run-clear-eval-analysis --config_path=<path_to_config>
\end{lstlisting}
\end{myshellbox}

% Or with a single CLI command:
Or with the next three lines of Python code:

\begin{mycodebox}
\begin{lstlisting}
from clear_eval.analysis_runner import run_clear_eval_analysis
config_path = <path-to-config>
run_clear_eval_analysis(config_path)
\end{lstlisting} % Include your file here
\end{mycodebox}

where <path\_to\_config> is the path to a YAML file containing the basic configuration options, such as the selected judge and provider, path to input data, path to the results folder, and other optional parameters.

Once processing is complete, the Streamlit interface can be launched using the CLI command below. The results ZIP file saved to the specified output directory can then be manually loaded through the app.

%the results are saved to the specified output directory and can be explored through our interactive Streamlit interface :

% Then, based on the saved result file, we can launch the Streamlit interference and load the target results file.

\begin{myshellbox}
\begin{lstlisting}
$ run-clear-eval-dashboard
\end{lstlisting} % Include your file here
\end{myshellbox}

\subsection{UI}
% Figure with main parts

% Filtering - and/or/not;
% Deep-dive - all the info instances, including the judge feedback, summary, and issues - allow for a better understanding of the way the issues are expressed.

To support intuitive and effective exploration of model errors, \package{} includes a visual analytics interface designed for both researchers and practitioners. The user interface (Figure~\ref{fig:ui}) provides multiple synchronized views that together offer a comprehensive understanding of model behavior, error patterns, and their distribution.

The interface is composed of the following key components:

% \paragraph{Opening view.}
% This panel (Figure~\ref{fig:ui}a) serves as the entry point to the analysis tool. It allows loading the evaluation dataset and provides basic statistics, such as data size and score distributions. Users can quickly grasp the dataset's scope and quality, and begin filtering or inspecting it with one click.

\paragraph{Issues View.}
The Issues View (Figure~\ref{fig:ui}b) displays an overview of all the issues identified by the system. Each issue is listed along with its frequency and percentage in the dataset. This helps users identify dominant failure patterns at a glance. It also helps the user understand the severity of the presented issues.
% Insights are shown alongside, e.g., indicating that mathematical errors are the most frequent failure type in a given evaluation.

\paragraph{Filtering Mechanism.}
The filtering panel (Figure~\ref{fig:ui}c) enables users to narrow down the dataset based on a specific combination of issue types or score range. We allow the union or intersection of issues or their negation. This is essential for targeted exploration—for example, isolating only high-scoring answers with logical errors, or filtering for responses that demonstrate extractiveness issues. The panel offers dynamic control, and any applied filters immediately update the rest of the views.

\paragraph{Comparison View.}
The Comparison View (Figure~\ref{fig:ui}d) visualizes how issue frequencies change when filtering is applied. This comparison allows users to better understand the connection between different issues and, issues and score range.
% how they are manifested in different score ranges.
% correlations between issues 
% for example, understanding that responses with logical errors frequently also contain mathematical mistakes. This is particularly useful when debugging models or comparing fine-tuned versions.

\paragraph{Model Behavior and Instance-Level View.}
The bottom row of the interface (Figure~\ref{fig:ui}e, f) focuses on instance-level analysis. Users can drill down into specific examples, inspect the original instruction and response, the judge’s textual feedback, and the list of issues it was mapped to. This direct link between abstract issues and concrete examples can help users understand how different issues effect their system behavior.
% the tool actionable for error diagnosis, dataset curation, and model improvement.

Together, these views make \package{} a tool for analyzing LLM output beyond scalar metrics. The interface supports both broad patterns and fine-grained inspection, helping practitioners uncover failure trends, identify brittle behavior, and better understand how model responses fail in practice.

% \section{Demonstrating \package{} Capabilities}

\section{\package{}: Case Study}

\subsection{Setup}
To study our method's behavior, we utilize three datasets: GSM8K~\cite{cobbe2021trainingverifierssolvemath} for math word problems, and two retrieval-augmented generation (RAG) datasets: TechQA~\cite{castelli2019techqadataset} and DelucionQA~\cite{sadat-etal-2023-delucionqa}, based on the processing of RAGBench~\cite{friel2024ragbench}. We run evaluations over four open systems: Mixtral 8x7B~\cite{jiang2024mixtralexperts}, LLaMA-3.1 8B~\cite{grattafiori2024llama3herdmodels}, Granite-3.3 8B~\cite{granite2024granite}, and Phi-4~\cite{abdin2024phi4technicalreport}.

We generate responses for each dataset using these systems and apply our pipeline to assess them. For the judgment component, we employ two strong models in a reference-less setting with general and task-specific modes. As a high-quality closed-source judge, we use GPT-4o~\cite{openai2024gpt4ocard}, and as an open-source alternative, we use LLaMA-3.3 70B. The per-instance feedback generated by these judges is passed through two versions of the $K$ module: the IBM watsonx\textsuperscript{\textregistered} Key Point Analysis (KPA) implementation\footnote{\href{https://www.ibm.com/docs/en/watsonx/saas?topic=scripts-key-point-summarization}{IBM watsonx KPA}}, and our LLM-Based KPA using both GPT-4o and LLaMA-3.3 70B.
This full pipeline produces, for each system–dataset pair, a set of recurring issues derived from the judge feedback.

% a tight itemize style we’ll use in the table
\newlist{tightitem}{itemize}{1}
\setlist[tightitem]{
  nosep,               % no extra vertical space
  leftmargin=0.2em,      % small indent
  label=--,            % short en-dash bullet (change as you like)
}
% \newcommand{\itemdash}{\vspace{0.3em}\hdashrule[0.5ex]{\linewidth}{0.4pt}{2pt}\vspace{0.3em}}

% ------------------------------------------------------

% --------------- the actual table ---------------------
\renewcommand{\arraystretch}{0.95}   % ← tiny row-height squeeze (optional)

\begin{table}[t!]
\centering
\small
\begin{tabularx}{\columnwidth}{>{\raggedright\arraybackslash}X}
\toprule
\multicolumn{1}{c}{\textbf{GSM8K}    ($s$: Mixtral 8x7B, $J$: GPT-4o)} \\
\midrule
\parbox{\linewidth}{
\begin{tightitem}
\item No Issues Detected ($78.4\%$)
  \item Mathematical errors in calculations, including rounding and final steps. ($13.2\%$)
  \item Incorrect understanding of problem statements leading to flawed reasoning. ($11.8\%$)
  \item Failure to fully consider or correctly interpret all given information. ($5.8\%$)
  \item Incomplete answers due to missing necessary steps or calculations. ($5.5\%$)
  \item Logical errors despite clear reasoning. ($4.3\%$)
  \item Misunderstanding or incorrect application of mathematical concepts or methods. ($3.3\%$)
  \item Unnecessary complexity or inclusion of irrelevant details. ($2.6\%$)
  \item Incorrect handling of units or conversions. ($0.6\%$)
  \item Failure to verify or cross-check results. ($0.2\%$)
\end{tightitem}
} \\
\bottomrule
\end{tabularx}
\caption{
Issues identified for Mixtral 8x7B over the GSM8K benchmark with a task-specific mode, sorted by decreasing frequency (shown in the parentheses).} %Analysis of model errors on different benchmarks (issues are sorted based on their frequency).}
\label{tab:gsm8k} %\vspace{-5pt}
\end{table}
\begin{table}[t!]
\centering
\small
\begin{tabularx}{\columnwidth}{>{\raggedright\arraybackslash}X}
\toprule
\multicolumn{1}{c}{\textbf{Mixtral 8x7B} ($J$: GPT-4o) } \\
\midrule
\parbox{\linewidth}{
\begin{tightitem}
\item No Issues Detected ($51.9\%$)
\item Omission of necessary details or steps ($36.3\%$)
\item Lack of specificity and completeness in responses ($31.2\%$)
\item Omission of relevant links or references ($9.2\%$)
\item Inaccurate or irrelevant information ($8.6\%$)
\item Failure to provide actionable insights or solutions ($8.3\%$).
\item Misinterpretation or misuse of context($4.5\%$)
\item Lack of clarity in explaining technical details ($3.5\%$)
\item Incomplete or abrupt ending of the response ($3.5\%$)
\end{tightitem}
} \\
\midrule
\multicolumn{1}{c}{\textbf{Phi-4} ($J$: GPT-4o) } \\
\midrule
\parbox{\linewidth}{
\begin{tightitem}
\item No Issues Detected ($76.6\%$)
\item Lacks completeness and necessary details ($10.9\%$)
\item Lacks context-specific information ($9.9\%$)
\item Lacks specificity in technical details ($6\%$)
\item Fails to mention unsupported features or limitations ($5.1\%$)
\item Inaccurate or fabricated information ($2.6\%$)
\item Does not directly answer the question ($1.9\%$)
\item Assumes unsupported or incorrect context ($1.9\%$)
\end{tightitem}
} \\
\bottomrule
\end{tabularx}
\caption{Issues identified for Mixtral 8x7B and Phi-4 over TechQA with a general mode, sorted by decreasing frequency (shown in the parenthesis)}
\label{tab:issues}  %\vspace{-8pt}
\end{table}

\subsection{Results}\label{sec:res}

% \paragraph{Task-and benchmark-specific issue discovery.} 

\paragraph{Actionable Insights.}
Table~\ref{tab:gsm8k} presents \package{} results of Mixtral 8x7B evaluated on GSM8K with a task-specific mode. The results reveal that Mixtral 8x7B most prominent weaknesses stem from calculation-related errors. Issues also include incorrect application of mathematical concepts and difficulties in handling units or conversions. 
These issues can help detect whether a model's failures stem from reasoning gaps, execution errors, or both.
Importantly, this insight is actionable: developers may choose to augment training data with synthetic examples focusing on numeric reasoning~\cite{yehudai2024genieachievinghumanparity}, while users might compensate by pairing the model with external tools like calculators~\cite{qin2023toolllmfacilitatinglargelanguage}.

% Table~\ref{tab:issues} presents results for two representative cases: Mixtral 8x7B evaluated on GSM8K with a task-specific mode, and Phi-4 on TechQA, with a general mode. Both were analyzed using GPT-4o as the judge and the LLM-based KPA.
% Table~\ref{tab:gsm8k} presents results f

% As presented in Table~\ref{tab:gsm8k}, on GSM8K, \package{} reveals that Mixtral 8x7B most prominent weaknesses stem from calculation-related errors. Issues also include incorrect application of mathematical concepts and difficulties in handling units or conversions. 
% These issues can help detect whether a model's failures stem from reasoning gaps, execution errors, or both.
% % Such issues indicate that the model’s failures arise not only from reasoning gaps but also from execution-level mistakes. 
% Importantly, this insight is actionable: developers may choose to augment training data with synthetic examples focusing on numeric reasoning, while users might compensate by pairing the model with external tools like calculators.

\paragraph{Data Dependent Issues.}
In the upper part of Table~\ref{tab:issues} we present the \package{} results from evaluating Mixtral 8x7B (with \textit{J}: GPT-4o) on TechQA with a general mode. Issues include missing context, lack of specificity in technical content, and hallucinated or fabricated information—a well-known failure mode of LLMs in RAG settings.
Notably, \package{} demonstrates the ability to adapt issue discovery to both the task and the dataset. For example, the extracted issues explicitly reflect TechQA’s demand for accurate, domain-specific technical details, illustrating how the system tailors feedback to the characteristics of each benchmark.

\paragraph{System Impact.} To study how \package{} behaves across different systems, we analyze results on TechQA for Mixtral 8x7B and Phi-4 (see Table~\ref{tab:issues}). Firstly, we observe that the two systems produce different sets of issues, highlighting that \package{} provides system-specific diagnostic feedback. For example, Mixtral 8x7B exhibits unique problems such as ``Omission of relevant links or references'' and ``Failure to provide actionable insights or solutions,'' which are not observed for Phi-4.

Secondly, we find that the overall proportion of flagged instances is lower for Phi-4, 48.1\% for Mixtral 8x7B versus 23.4\% for Phi-4. This aligns with the models’ overall quality as reported in the literature: Phi-4 achieves an 84.8 MMLU score~\cite{hendrycks2021measuringmassivemultitasklanguage} and a 1257 Elo score on Chatbot Arena~\cite{chiang2024chatbotarenaopenplatform}, compared to Mixtral 8x7B’s 1194 MMLU and 1194 Elo scores. These findings suggest that \package{} can support high-level system comparisons, in addition to offering fine-grained diagnostic feedback.

% \paragraph{System Impact} To study how \package{} output behaves for different systems, we examine the results over TechQA for Mixtral 8x7B, and Phi-4 (See Table~\ref{tab:issues}). Firstly, we can recognize that the proportion of instances flagged with issues is lower for Phi-4, e.g.\lil{i.e.?}, $48.1\%$ and $23.4\%$ issues for Mixtral 8x7B, and Phi-4, respectively. 
% This is in line with the model's overall quality as known from the literature, Phi-4 scores 84.8 on MMLU \cite{hendrycks2021measuringmassivemultitasklanguage}, and has a 1257 Elo score on Chatbot
% Arena \cite{chiang2024chatbotarenaopenplatform}, compared to Mixtral 8x7B with an MMLU score of 1194, and an Elo of 1194.
% This suggests that \package{} can also facilitate overall system comparison, besides the more fine-grained feedback.

% Secondly, we observe that the two systems produce different sets of issues, highlighting that \package{} provides system-specific diagnostic feedback. For example, Mixtral 8x7B suffers from unique issues like ``Omission of relevant links or references'', or ``Failure to provide actionable insights or solutions'' which are not present for Phi-4.

% See Appendix~\ref{} for the complete list of issues.

\paragraph{Impact of Evaluation Modes.}
\label{evaluation_mode}
% We also compare the use of task-specific and general evaluation modes.
% Our findings show that task-specific mode increases the sensitivity of \package{} to capture the task-specific issues. In contrast, the general mode often yields a broader variety of more nuanced issues. 
% For example, on RAG datasets, task-specific mode helped expose more faithfulness-related issues, which were overlooked by the general prompt, e.g. ``Generates unsupported or speculative information''. On the other hand, the general prompt discovered more novel issues, such as ``Omission of relevant links or references'' and ``Incomplete or abrupt ending of the response''(Appendix~\ref{app:prompt_selection}).

We also compare the use of task-specific and general evaluation modes.
Our findings show that the task-specific mode increases \package{}'s sensitivity to issues closely tied to the task. In contrast, the general mode tends to reveal a broader range of more nuanced or unanticipated problems.
For example, on RAG datasets, the task-specific prompt helped to expose more faithfulness-related issues, such as ``Generates unsupported or speculative information'', that were missed by the general prompt. On the other hand, the general mode discovered more novel issues like ``Incomplete or abrupt ending of the response''(See appendix~\ref{app:prompt_selection} for a full comparison).

% \mss{this is from the table above of the mixtral. Not clear to me how in the Data dependent issues it was focused on rag metrics..}

\paragraph{Impact of KPA method.}
Finally, we quantitatively assess the three implementations of the $K$ module for key point analysis. Our results indicate that LLM-based KPA tends to produce issues that are more synthesized and less extractive than the traditional KPA approach. Among the models evaluated, GPT-4o produced more accurate and specific issue types compared to LLaMA-3.3 and Watsonx’s implementation (see Appendix~\ref{app:compar}).

\subsection{User study}
% User study   - useful UI

To evaluate the usefulness and usability of \package{} for users, we conducted a user study with 12 AI practitioners and researchers. Participants were asked to use the tool on the three datasets, explore the interface, and provide feedback via a structured questionnaire (On a Likert scale) and free-form comments. The assessed dimensions included the usefulness of the tool, comparison to their current practices, and their trust in the tool. (see App.~\ref{app:user_study} for a detailed description of the participants, instructions, and questions).

% \paragraph{Key findings.}
% The results indicate that users found \package{} valuable for surface-level analysis, saving time, and offering a more accessible and interactive experience than their current manual workflows. 
% Many highlighted the ability to quickly identify and categorize errors, drill down into examples, and gain actionable insights. Several users noted that the system made it easier to spend time on error analysis and to uncover failure patterns they might have otherwise missed.

\paragraph{Results.}
The results indicate that users found \package{} valuable for surface-level analysis. Specifically, participants appreciated the automation of error detection (noting that $75\%$ currently rely on manual inspection for their use cases), the visual exploration interface, and the potential to detect issues they would have overlooked, with an average rating of $4.33$ on a Likert scale.
The system was seen as \textit{actionable}, with $74\%$ of participants reporting they would take or consider taking action based on the output, \textit{time-saving}, and \textit{better than existing practices}, with both aspects receiving average scores of 4.25. 
% The system was seen as \textit{actionable} (41\% reported they would consider taking action based on the output, 33\% indicated they would take action, and the remainder were unsure), \textit{time-saving}, and \textit{better than existing practices}, with both aspects receiving average scores of 4.25. 
It was especially valued for identifying common failure modes at scale, with an average score of 4.16. 

% \paragraph{Limitations and Opportunities.}
Despite the generally positive reception, users pointed out areas for improvement. Several responses expressed uncertainty regarding the \textit{trustworthiness}, $3.83$ score, and \textit{specificity} of the surfaced issues. Some found the descriptions to be vague or had difficulty understanding which errors were most critical. Users also requested features such as severity annotations, clearer categories, automatic summaries, and better highlighting within the textual feedback.

\section{Related Work}

Existing error analysis tools map model errors and abilities by inspecting dataset instances, such as EvalTree's \cite{zeng2025evaltree} capability hierarchies or Qualeval's \cite{murahari2023qualeval} data attributes. Other methods are interactive, like Erudite \cite{wu2019errudite}, which requires user labels for clustering errors, or use specialized models like MisattributionLLM \cite{xu2025misattributionllm} to score known error types.

Crucially, all these methods depend on labeled data, restricting them to specific tasks. Moreover, as these works probe for weaknesses or skills based on dataset features rather than model-specific behavior, they are likely to miss idiosyncratic model failure modes.

\section{Conclusion and Future work}

% \mss{this is the conclusion from the user study.}
% Overall, the study shows that \package{} is a promising tool for accelerating and structuring error analysis workflows. It performs well in surfacing surface-level behaviors and facilitating actionable insights. However, there remains room to enhance its interpretability, trust calibration, and depth of analysis—important areas for future work.

We presented \package{}, a novel framework and interactive tool for automating error analysis of generative AI systems. By leveraging LLMaJ evaluations and Key Point Analysis, \package{} extracts structured, system-level feedback from instance-level critiques. This enables AI developers to move beyond scalar metrics and surface recurring, actionable failure patterns with minimal manual overhead.

Our experiments across math and RAG datasets demonstrate that \package{} adapts to different tasks and models, revealing both common and system-specific issues. The tool offers flexibility in evaluation modes, supports multiple judges and KPA configurations, and includes an intuitive visual interface for deep exploration of model behavior. Our user study confirms the tool’s value for practitioners, highlighting its ability to save time, provide new insights, and improve analysis workflows, though it also points to areas for further enhancement.

Looking ahead, we plan to improve the specificity and clarity of the discovered issues, incorporate severity scoring and prioritization mechanisms, and explore methods for increasing user trust and interpretability. Additionally, we aim to integrate interactive feedback loops that allow users to refine or correct discovered issues.

\package{} is publicly available, and we hope it will serve the community as a stepping stone toward more transparent, efficient, and insightful evaluation of generative models.

\bibliography{custom}

\appendix
\newpage

\section{Limitations}

Our framework's effectiveness is fundamentally dependent on the quality of its constituent models. The analysis pipeline inherits the biases and potential inaccuracies of both the underlying LLM-as-a-Judge ($J$) and the Key Point Analysis ($K$) module.
\paragraph{Dependence on Judge Quality} The discovered issues are only as reliable as the initial critiques from the judge model. Biases in the judge (e.g., self-bias, length/style bias) or their failure to identify subtle errors directly compromise the quality and validity of the final, aggregated issues.

\paragraph{Scalability and Cost} The LLM-based KPA approach, while effective, can be computationally expensive, requiring approximately $\sim2N$ LLM calls for a dataset of size $N$. 
However, we apply the LLM-based KPA only to instances with low initial evaluation scores, making the cost dependent on the target system’s quality. Nevertheless, this overhead may present a practical limitation for large-scale analyses.

\paragraph{Lack of Causality} Our tool identifies and quantifies recurring error patterns but does not diagnose their root cause. For instance, it can highlight frequent factual inaccuracies but cannot distinguish if this stems from knowledge deficits, retrieval failures, or flawed reasoning.

\section{Method Comparison}
\label{app:compar}

%  To add some details about the systems and the data
Table~\ref{tab:model_eval_grid_final} presents a qualitative comparison of the key points generated by the three KPA implementations: Watsonx-KPA, LLM-based KPA with LLaMA-3, and LLM-based KPA with GPT-4o. While all three aim to identify recurring system-level issues, they differ in terms of abstraction level, phrasing style, and the generality of the resulting key points. Below, we describe two central dimensions of variation we observed across the methods.

\paragraph{Extractive vs. Synthesized Styles.}
The Watsonx-KPA method produces highly extractive and granular key points, often lifted nearly verbatim from the original feedback. As a result, its issues tend to be overly specific (e.g., \textit{``The calculation in step 7 is incorrect''}) or tied to particular instance structures. While this precision can surface concrete issues, it reduces generalizability and often results in a long list of narrowly scoped problems. In contrast, LLM-based KPA—especially with GPT-4o—generates more abstracted and synthesized issue types. These key points aggregate multiple occurrences of related errors into broader categories, such as \textit{``Failure to fully consider or correctly interpret all given information''}, which promote better generalization across examples and systems.

\paragraph{Issue Granularity and Clarity.}
LLM-based methods also differ in their balance between abstraction and clarity. The LLaMA-3-based KPA sometimes produces long, compound key points with broad scopes (e.g., \textit{``Failure to account for all relevant variables, conditions, or scenarios''}), which may reduce readability and introduce redundancy. GPT-4o, on the other hand, tends to produce concise and well-structured key points, striking a balance between informativeness and clarity (e.g., \textit{``Incorrect handling of units or conversions''}). In contrast, Watsonx’s output can feel fragmented or repetitive due to its instance-tethered phrasing, often leading to multiple key points covering overlapping aspects of the same underlying issue.

\section{Impact of evaluation mode}
\label{app:prompt_selection}

\begin{table*}[t!]
\centering
\small
\begin{tabularx}{\textwidth}{l >{\raggedright\arraybackslash}X >{\raggedright\arraybackslash}X >{\raggedright\arraybackslash}X}
\toprule
\textbf{Dataset} & \textbf{watsonx-KPA}  & \textbf{LLM-based (GPT-4o)} & \textbf{LLM-based (LLaMA-3-70B)} \\
\midrule

\textbf{GSM8K} & 
%\parbox{\linewidth}{
\begin{tightitem}
\item No Issues Detected ($78.3\%$)
  \item The error leads to an incorrect final answer. ($17.8\%$)
\item The equations are set up inaccurately. ($14.4\%$)
\item The model fails to provide correct reasoning. ($12.3\%$)
\item The calculation in step 7 is incorrect. ($11.2\%$)
\item The error leads to an incorrect average. ($10.5\%$)
\item However, the rounding error could be misleading. ($7.1\%$)
\item It also lacks clarity. ($6.3\%$)
\item It does not verify calculations. ($5.6\%$)
\item The steps in the model answer are incomplete. ($5.4\%$)
\item The model incorrectly calculates the number of pairs. ($4.5\%$)
\item Necessary steps to solve the problem are missing. ($3.9\%$)
\end{tightitem}
%} 
& 
%\parbox{\linewidth}{
\begin{tightitem}
 \item No Issues Detected ($83.7\%$)
\item Incorrect understanding of problem statements leading to flawed reasoning ($7.5\%$)
\item Mathematical errors in calculations, including rounding and final steps ($7.4\%$)
\item Incomplete answers due to missing necessary steps or calculations ($4.1\%$)
\item Failure to fully consider or correctly interpret all given information ($3.9\%$)
\item Unnecessary complexity or inclusion of irrelevant details ($1.7\%$)
\item Logical errors despite clear reasoning ($1.1\%$)
\item Incorrect handling of units or conversions ($0.4\%$)
\item Misunderstanding or incorrect application of mathematical concepts or methods ($0.3\%$)
\item Failure to verify or cross-check results ($0.2\%$)
\end{tightitem}
%}
& 
%\parbox{\linewidth}{
\begin{tightitem}
  \item No Issues Detected (81.4%)
\item Calculation errors or inaccuracies ($15.2\%$)
\item Flawed reasoning, logical errors, or incorrect application of formulas/algorithms ($7.1\%$)
\item Incorrect assumptions or misinterpretations of problem statements ($5.2\%$)
\item Failure to account for all relevant variables, conditions, or scenarios ($5.2\%$)
\item Lack of clarity, consistency, or unnecessary complexity in explanations ($3.1\%$)
\item Inability to correctly interpret or apply given information ($2.9\%$)
\end{tightitem}
% }
\\

\midrule

\textbf{TechQA} & 
%\parbox{\linewidth}{
\begin{tightitem}
\item No Issues Detected ($4.5\%$)
\item The response lacks completeness and clarity. ($89.5\%$)
\item The model answer lacks relevance and factual support. ($70.7\%$)
\item It fails to provide document-supported solutions. ($59.2\%$)
\item The model misses key details about the vulnerability. ($55.1\%$)
\item As a result, the response lacks faithfulness. ($54.8\%$)
\item It contains unnecessary details and inaccuracies. ($41.4\%$)
\item It misses several necessary troubleshooting steps. ($26.4\%$)
\item The model lacks actionable steps and insights. ($21.3\%$)
\item However, it lacks specific configuration steps for ITCAM. ($8.6\%$)
\item However, the additional join information is unnecessary. ($6.4\%$)
\item The answer misses specific details about cache synchronization. ($3.8\%$)
\end{tightitem}
%}
& 
%\parbox{\linewidth}{
\begin{tightitem}
  \item No Issues Detected ($18.5\%$)
\item Lacks completeness and omits crucial details ($59.6\%$)
\item Generates unsupported or speculative information ($31.8\%$)
\item Fails to accurately incorporate document information ($22.0\%$)
 \item Provides irrelevant or extraneous information ($14.3\%$)
 \item Lacks clarity and conciseness ($14.0\%$)
 \item Fails to address the specific question ($12.4\%$)
 \item Fails to provide direct solutions ($8.0\%$)
 \item Lacks structured presentation ($3.8\%$)
\end{tightitem}
%} 
& 
%\parbox{\linewidth}{
\begin{tightitem}
 \item No Issues Detected 26 ($8.3\%$)
\item Lack of detail and clarity in answers ($67.8\%$)
\item Inadequate consideration of context and user needs ($45.9\%$)
\item Failure to directly address the user's question ($43.3\%$)
\item Unclear or incomplete solutions ($36.6\%$)
\item Lack of faithfulness to original documents ($26.8\%$)
\item Introduction of unnecessary information ($24.2\%$)
\item Failure to provide supporting evidence or examples ($22.3\%$)
\item Failure to provide clear steps or instructions ($19.7\%$)
\item Lack of relevance to the specific question or topic ($19.4\%$)
\item Inaccuracies and inconsistencies in information ($15.3\%$)
\item Overreliance on general knowledge ($14.0\%$)
\item Failure to consider multiple possible causes or solutions ($8.3\%$)
\item Inadequate explanation of technical terms ($2.2\%$)
\end{tightitem}
%}
\\

\bottomrule
\end{tabularx}
\caption{Top issues identified for Mixtral 8x7B, with each method across GSM8K and TechQA benchmarks, with a task-specific evaluation mode.}
\label{tab:model_eval_grid_final}
\end{table*}

Table~\ref{tab:eval_mode} shows the full list of discovered issues for mixtral 8x7b over the TechQA benchmark, using the general vs. the task-specific evaluation mode, as discussed in paragraph \textit{Impact of Evaluation Modes} under section
~\S\ref{sec:res}.

\begin{table}[t!]
\centering
\small
\begin{tabularx}{\columnwidth}{>{\raggedright\arraybackslash}X}
\toprule
\multicolumn{1}{c}{\textbf{General} ($J$: GPT-4o) } \\
\midrule
\begin{tightitem}
\item No Issues Detected ($51.9\%$)
\item Omission of necessary details or steps ($36.3\%$)
\item Lack of specificity and completeness in responses ($31.2\%$)
\item Omission of relevant links or references ($9.2\%$)
\item Inaccurate or irrelevant information ($8.6\%$)
\item Failure to provide actionable insights or solutions ($8.3\%$)
\item Misinterpretation or misuse of context($4.5\%$)
\item Lack of clarity in explaining technical details ($3.5\%$)
\item Incomplete or abrupt ending of the response ($3.5\%$)
\end{tightitem}
\\
\midrule
\multicolumn{1}{c}{\textbf{Task-Specific} ($J$: GPT-4o) } \\
\midrule
\begin{tightitem}
\item No Issues Detected ($18.5\%$)
\item Lacks completeness and omits crucial details ($59.6\%$)
\item Generates unsupported or speculative information ($31.8\%$)
\item Fails to accurately incorporate document information ($22.0\%$)
 \item Provides irrelevant or extraneous information ($14.3\%$)
 \item Lacks clarity and conciseness ($14.0\%$)
 \item Fails to address the specific question ($12.4\%$)
 \item Fails to provide direct solutions ($8.0\%$)
 \item Lacks structured presentation ($3.8\%$)
\end{tightitem}
\\
\bottomrule
\end{tabularx}
\caption{Issues identified for Mixtral 8x7B over TechQA, using general (top) and task-specific (bottom) evaluation modes.}
\label{tab:eval_mode}
\end{table}

\section{User Study Details}
\label{app:user_study}

Among the 12 participants, 7 are application developers, 3 are business analysts, and 1 is a model developer.

\begin{figure}
\includegraphics[width=0.45\textwidth]{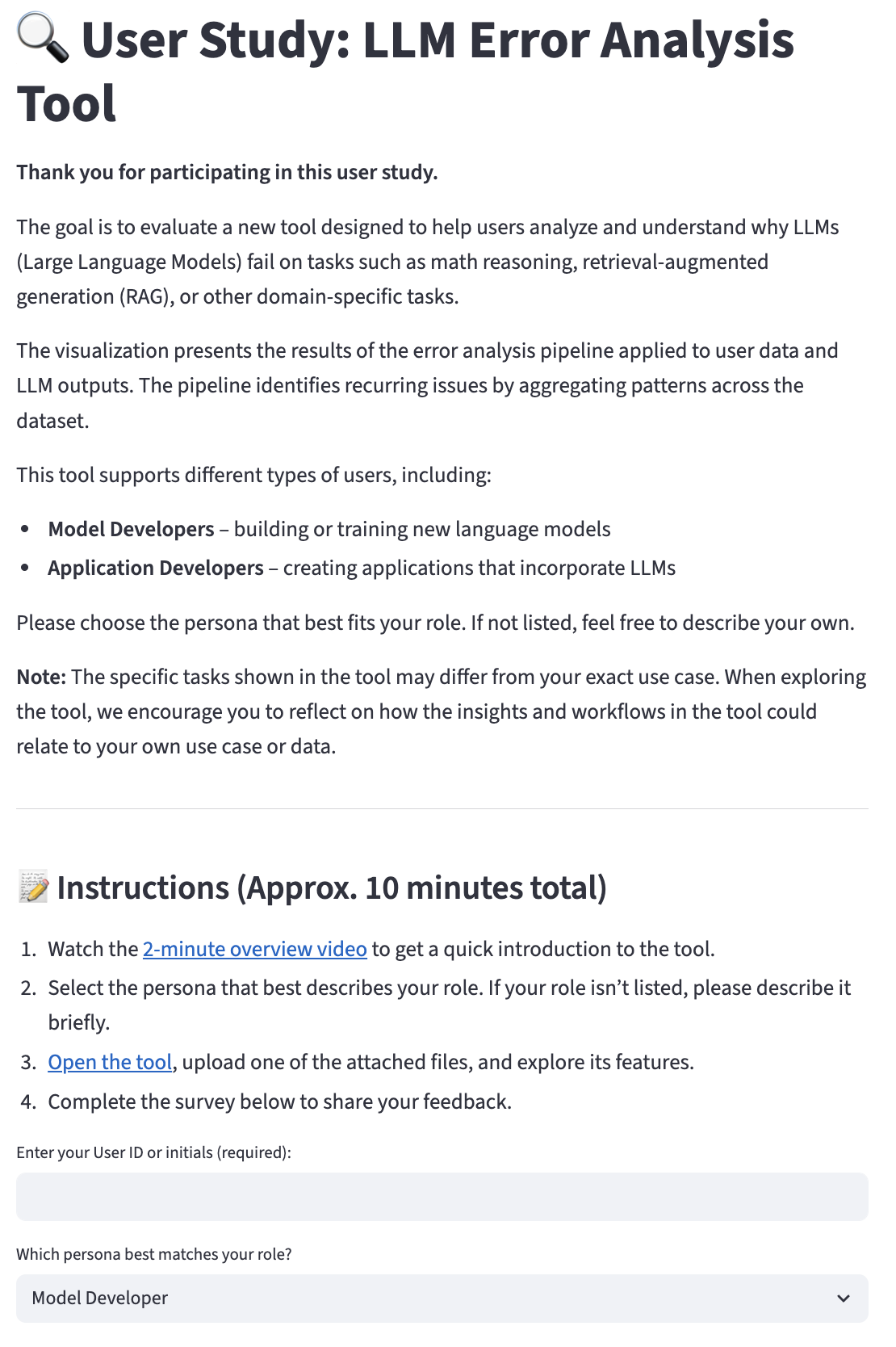}
\caption{Instructions to the study participants.}
\label{fig:study_inst}
\end{figure}

To set the context for the study, we began with a brief overview of the tool, followed by step-by-step instructions (see Figure~\ref{fig:study_inst}).

\begin{figure}
\includegraphics[width=0.45\textwidth]{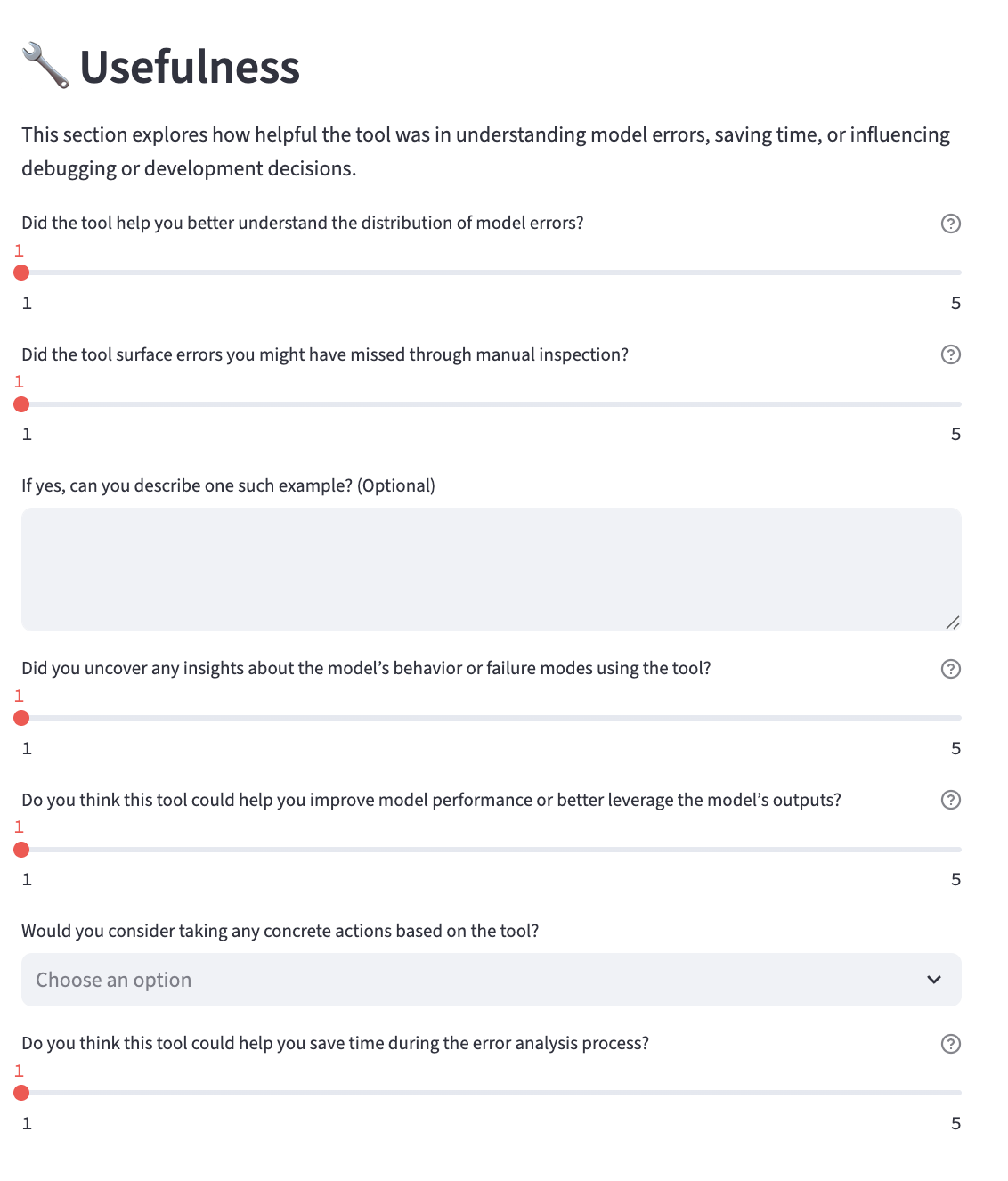}
\caption{Section 1- Usefulness questions.}
\label{fig:study_use}
\end{figure}

\begin{figure}
\includegraphics[width=0.45\textwidth]{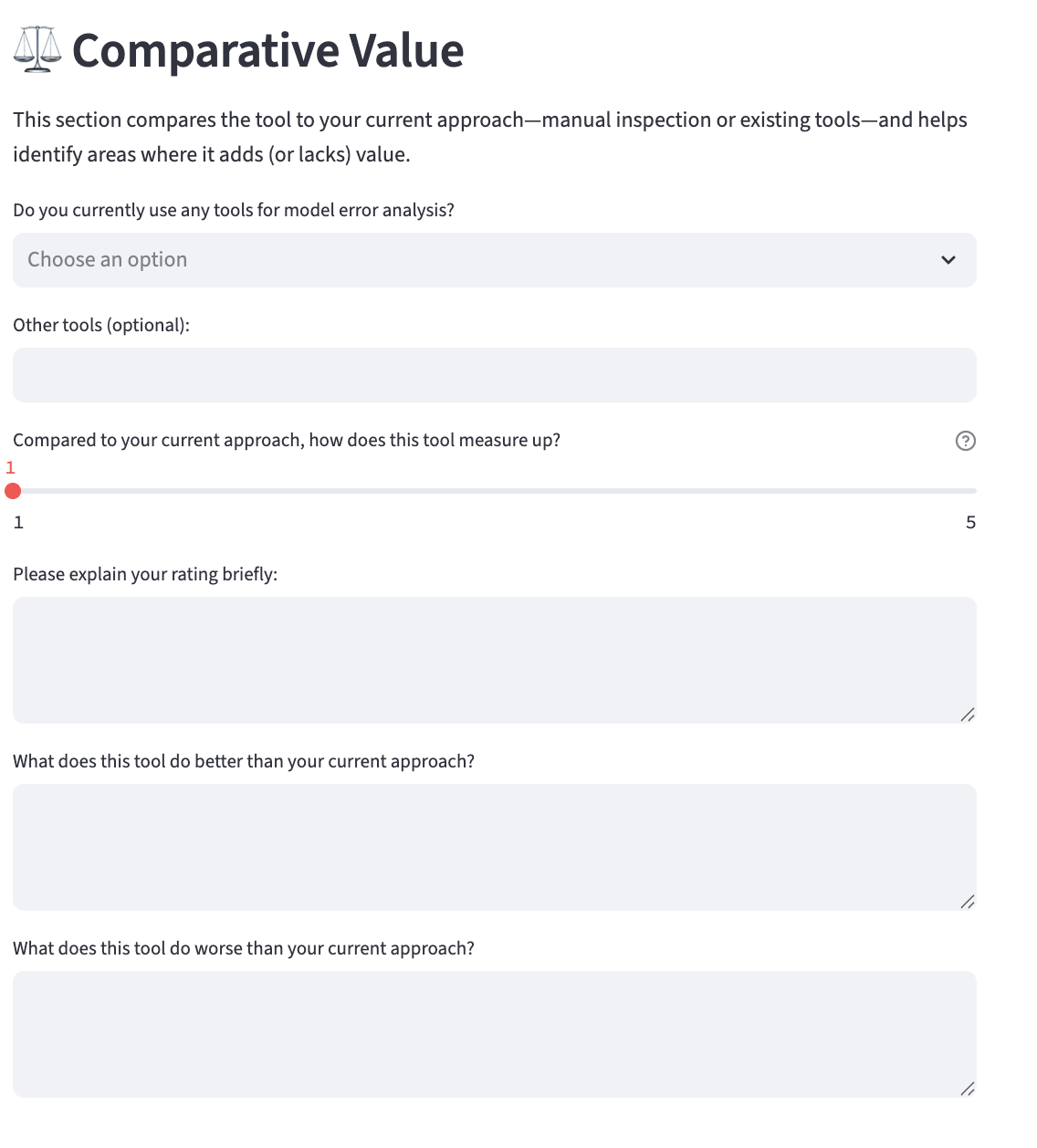}
\caption{Section 2- Comparative value questions.}
\label{fig:study_compare}
\end{figure}

\begin{figure}
\includegraphics[width=0.45\textwidth]{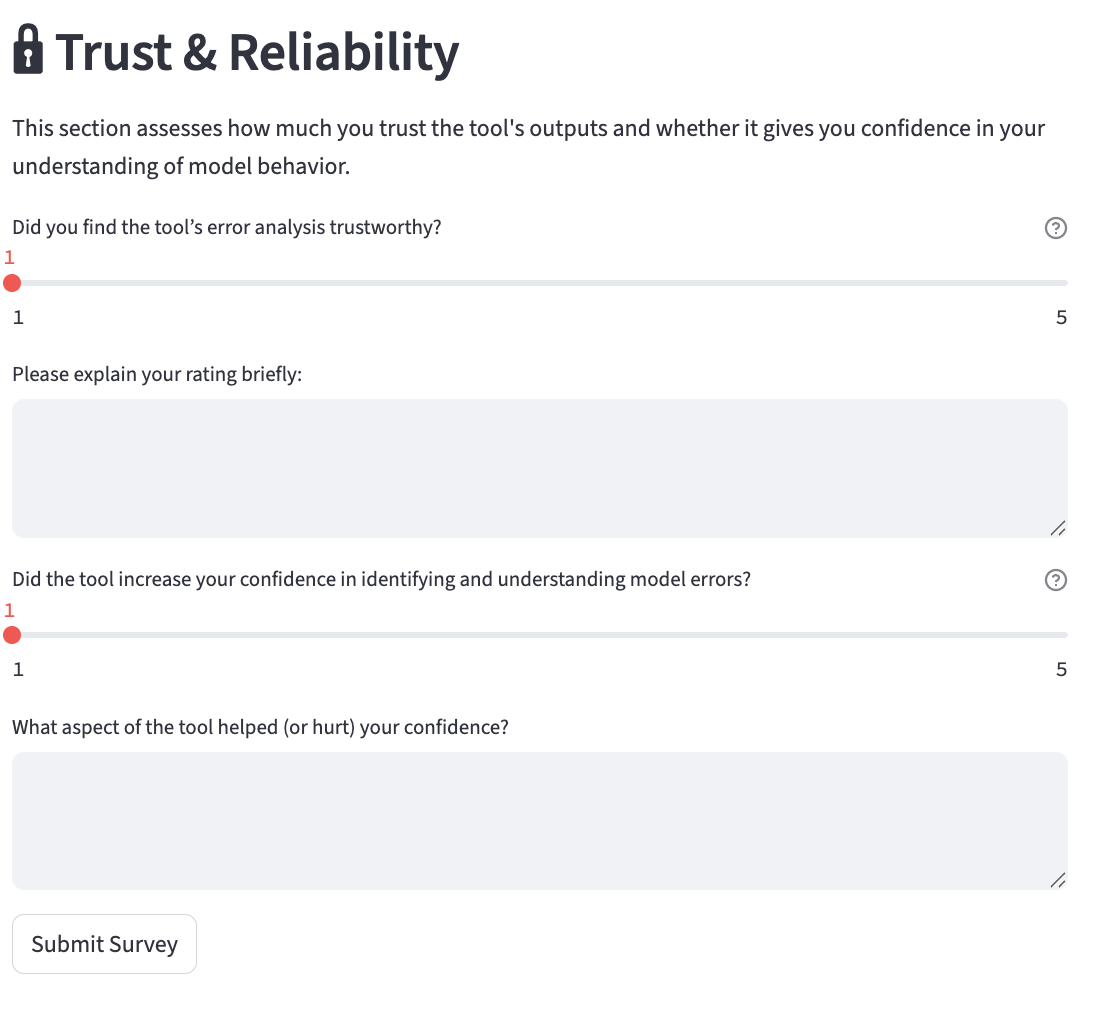}
\caption{Section 3- Trust \& Reliability questions.}
\label{fig:study_trust}
\end{figure}

The questions included a 1-5 Likert-scale, multiple-choice, and open-ended formats, divided into three sections to assess different aspects of the tool: 
\begin{enumerate}
    \item Usefulness - This section explores how helpful the tool was in understanding model errors, saving time, or influencing debugging or development decisions (Figure~\ref{fig:study_use}).
    \item Comparative value - This section compares the tool to your current approach—manual inspection or existing tools—and helps identify areas where it adds (or lacks) value (Figure~\ref{fig:study_compare}).
    \item Trust \& reliability- This section assesses how much you trust the tool's outputs and whether it gives you confidence in your understanding of model behavior (Figure~\ref{fig:study_trust}).
\end{enumerate}

% $75\%$ of the participants are doing manual inspections for their use-cases, and $25\%$ have their custom scripts or notebooks.
% Users' trust in the tool was medium-high, with an average score of $3.92$ on a 1-5 Likert scale.

\section{Implementation Details}
\label{app:promts}
% \paragraph{Prompt}
% The general evaluation prompt used by the judge to evaluate single instances is given in Fig~\ref{fig:general_eval_prompt}. The specific prompts used for the Math and RAG use cases are given in Fig~\ref{fig:math_eval_prompt} and ~\ref{fig:rag_eval_prompt}, respectively. Fig~\ref{fig:summary_prompt} presents the prompt for the summarization of each evaluation step. Fig~\ref{fig:synthesis}, \ref{fig:clustering}, and ~\ref{fig:mapping} show the prompts for issues list synthesis and deduplication. Finally, Fig~\ref{fig:mapping} shows the prompt used for mapping each instance to the matching issues.

\paragraph{Setup} For response generations, all models were prompted with default parameters. For all evaluation stages, inference was performed with temperature 0. The system was set to produce between 3 and 15 key points for each analysis. Issue synthesis was performed using up to 150 evaluation summaries with non perfect scores. All prompts are provided in the Git Repo.

% \input{prompts/general_eval}

% \input{prompts/math_eval}

% \input{prompts/rag_eval}

% \insertprompt{prompts/general_eval.txt}{Prompt: General Evaluation}{
% Prompt shown to the LLM judge in the general evaluation setting.
% }{fig:general_eval_prompt}

% \insertprompt{prompts/math_eval.txt}{Prompt: Math Evaluation}{
% Specific prompt shown to the LLM judge in the math evaluation setting.
% }{fig:math_eval_prompt}

% \insertprompt{prompts/rag_eval.txt}{Prompt: RAG Evaluation}{
% Specific prompt shown to the LLM judge in the RAG evaluation setting.
% }{fig:rag_eval_prompt}

% \insertprompt{prompts/summary.txt}{Prompt: Summarize Evaluation Text}{
% Prompt shown to the LLM judge for summarizing its evaluation text.
% }{fig:summary_prompt}

% \insertprompt{prompts/synthesis.txt}{Prompt: Issues Synthesis}{
% Prompt for synthesizing the recurring issues list.
% }{fig:synthesis}

% \insertprompt{prompts/clustering.txt}{Prompt: Issues Deduplication}{
% Prompt for consolidating overlapping issues into a distinct, merged list.
% }{fig:clustering}

% \insertprompt{prompts/mapping.txt}{Prompt: Mapping Records to Issues }{
% Prompt for mapping .
% }{fig:mapping}

\end{document}